\documentclass[10pt,twocolumn,letterpaper]{article}

\usepackage{cvpr}
\usepackage{times}
\usepackage{epsfig}
\usepackage{epstopdf}
\usepackage{graphicx}
\usepackage{amsmath}
\usepackage{amssymb}
\usepackage{times}  %Required
\usepackage{helvet}  %Required
\usepackage{courier}  %Required
\usepackage{url}  %Required
\usepackage{graphicx}  %Require

\usepackage{adjustbox}
\usepackage{amsmath}
\usepackage{color}
\usepackage{algorithm}
\usepackage{algorithmic}
\usepackage{array}
\usepackage{bm}
\usepackage{makecell,multirow,diagbox}

\usepackage[normalem]{ulem}

\usepackage[pagebackref=true,breaklinks=true,letterpaper=true,colorlinks,bookmarks=false]{hyperref}

\cvprfinalcopy % *** Uncomment this line for the final submission

\def\cvprPaperID{3145} % *** Enter the CVPR Paper ID here

\ifcvprfinal\fi
\begin{document}

%%%%%%%%% TITLE
\title{Towards Human-Machine Cooperation: \\ Self-supervised Sample Mining for Object Detection}

\author{Keze Wang$^{1,2}$ \quad\quad Xiaopeng Yan$^1$ \quad\quad Dongyu Zhang$^{1}\thanks{Corresponding author is Dongyu Zhang.}$ \quad\quad Lei Zhang$^{2}$ \quad\quad Liang Lin$^{1,3}$\\
{$^1$Sun Yat-sen University}~~~\ \
{$^2$The Hong Kong Polytechnic University}~~~\ \
{$^3$SenseTime Research}\\
{\tt\small kezewang@gmail.com; yanxp3@mail2.sysu.edu.cn; zhangdy27@mail.sysu.edu.cn}\\
{\tt\small cslzhang@comp.polyu.edu.hk; linliang@ieee.org}
%\and
}

\maketitle
\thispagestyle{empty}

%to boost object detection performance

%%%%%%%%% ABSTRACT
\begin{abstract}
Though quite challenging, leveraging large-scale unlabeled or partially labeled images in a cost-effective way has increasingly attracted interests for its great importance to computer vision. To tackle this problem, many Active Learning (AL) methods have been developed. However, these methods mainly define their sample selection criteria within a single image context, leading to the suboptimal robustness and impractical solution for large-scale object detection. In this paper, aiming to remedy the drawbacks of existing AL methods, we present a principled Self-supervised Sample Mining (SSM) process accounting for the real challenges in object detection. Specifically, our SSM process concentrates on automatically discovering and pseudo-labeling reliable region proposals for enhancing the object detector via the introduced cross image validation, i.e., pasting these proposals into different labeled images to comprehensively measure their values under different image contexts. By resorting to the SSM process, we propose a new AL framework for gradually incorporating unlabeled or partially labeled data into the model learning while minimizing the annotating effort of users. Extensive experiments on two public benchmarks clearly demonstrate our proposed framework can achieve the comparable performance to the state-of-the-art methods with significantly fewer annotations.
\end{abstract}

%%%%%%%%% BODY TEXT
 \section{Introduction}
In the past decade, object detection has gained incredible improvements both in accuracy and efficiency, benefiting from the remarkable success of deep Convolutional Neural Nets (CNNs)~\cite{alexnet12NIPS}\cite{googlenet}\cite{He_2016_CVPR}. Through producing candidate object regions of input images, object detection is converted into the region classification task, e.g., R-CNN~\cite{rcnn14CVPR}. Recently, more powerful neural network architectures such as ResNet~\cite{He_2016_CVPR} have further pushed the object detection performance into new records. Behind these successes, massive data collection and annotation such as MS-COCO~\cite{coco14ECCV} are indispensable yet quite expensive. Under such a circumstance, there is an increasing demand of leveraging large-scale unlabeled data to promote the detection performance in an incremental learning manner. However, to achieve this goal, there remain two technical issues: i) Object annotation for training is usually labor-intensive. Compared with other visual recognition task (e.g., image classification), annotating object requests  to provide both the category label and bounding box of an object. In order to reduce the burden of active users, it is highly required to develop human-machine cooperation based approaches to self-annotate most of the unlabeled data; ii) Picking out the training samples that are advantageous to boost the detection performance is a non-trivial task. As figured out in~\cite{ohem2016cvpr,hard16CVPR}, existing detection benchmarks usually contain an overwhelming number of ``easy'' examples and a small number of ``hard'' ones (i.e., informative samples with various illuminations, deformations, occlusions and other intra-class variations). Utilizing these ``hard'' samples is a key to train the model more effectively and efficiently. However, as pointed out in~\cite{Wang_2017_CVPR}, due to following a long-tail distribution, these examples are quite uncommon. Hence, it is a sophisticated task to find ``hard'' yet informative samples.

\begin{figure*}[t]
\center
\includegraphics[width = 1 \textwidth]{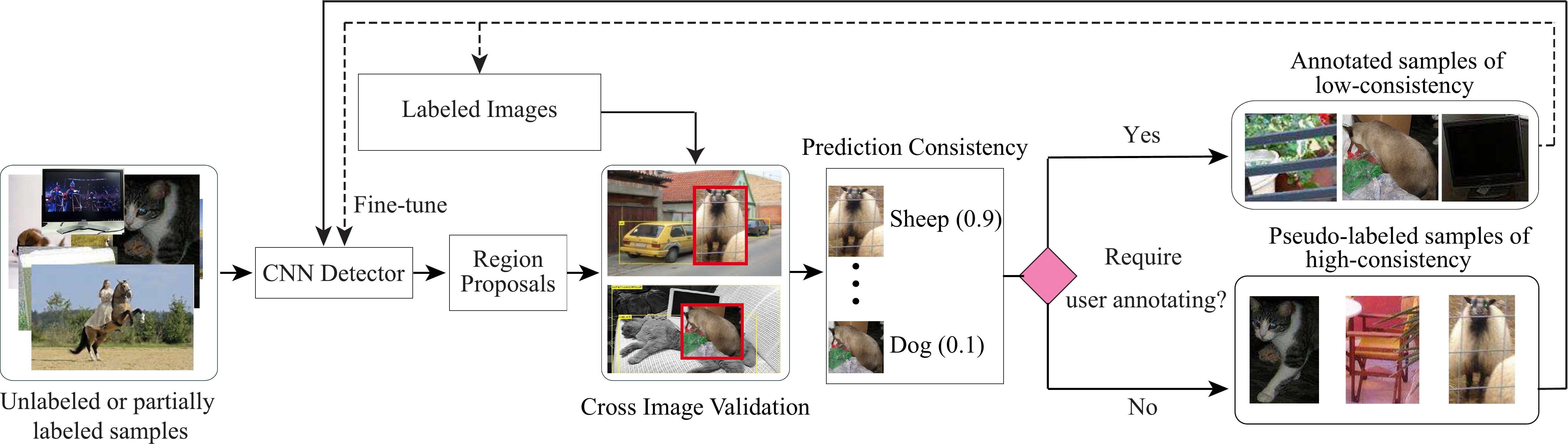}
\caption{The pipeline of the proposed framework with Self-supervised Sample Mining (SSM) process for object detection. Our framework includes stages of high-consistency sample pseudo-labeling via the SSM and low-consistency sample selecting via the AL, where the arrows represent the work-flow, the full lines denote data flow in each mini-batch based training iteration, and the dash lines represent data are processed intermittently. As shown, our framework presents a rational pipeline for improve object detection from unlabeled and partially labeled images by automatically distinguishing high-consistency region proposals, which can be easily and faithfully recognized by computers after the cross image validation, and low-consistency ones, which can be labeled by active users in an interactive manner.}
\label{fig:architecture}
\vspace{-8pt}
\end{figure*}

%feed heavy and computable tasks into computers, and arrange labor-saving and intelligent tasks to humans
%\cite{lewis1994sequential,tong2002support,joshi2009multi,Branson2011ICCV,Vijayanarasimhan2014}

To address the aforementioned issues, we investigate sample mining techniques to incrementally improve object detectors with minimal user effort. Recently, Active Learning (AL) \cite{ISed,lewis1994sequential,survey,bcip14CVPR,pose17ICCV} proposed to progressively select and annotate most informative unlabeled samples for user annotation to boost the model. Hence, the sample selection criteria play a crucial role in AL, and are typically defined according to the prediction certainty (confidence) or other informative criteria like diversity of samples. Recently, many AL methods~\cite{ceal16tcsvt, aspl17TPAMI,llal11CVPR} have been developed for training deep convolutional neural networks (CNNs). However, their sample selection criteria are dominantly performed under a single sample context. This make them sensitive to the bias of classifiers and the unbalance of samples.

Attempting to overcome the above-mentioned drawback of the existing AL approaches, this paper develops a Self-supervised Sample Mining (SSM) process for automatically mining valuable samples in terms of boosting performance of object detectors. Our developed SSM process is motivated by the recently popular self-supervised learning~\cite{Ghosh93self-organizationfor, MT_2017_ICCV,xl_2015_ICCV} technique. In stead of designing to learn a optimal visual representation as~\cite{xl_2015_ICCV,MT_2017_ICCV}, we concentrate on developing a rational pipeline of using the self-supervision to significantly decrease the amount of user annotations for improving detection performance. In the proposed SSM process, given the region proposals from unlabeled or partially labeled images, we evaluate their estimation consistency by performing the cross image validation, i.e., pasting them into different annotated images to validate its prediction consistency by the up-to-date object detector. Note that, to avoid ambiguity, the images for validation are randomly picked out from the labeled samples that do not belong to the estimated category of the under-processing proposal. Through a simple ranking mechanism, those incorrectly pseudo-labeled proposals have a large chance to be filtered due to the challenges inside various image contexts. In this way, the bias of classifiers and unbalance of samples can be effectively alleviated. Then, we propose to provisionally assign disposable pseudo-labels to the ones with high estimation consistency, and retrain the detectors within each mini-batch iteration. Since the pseudo-annotations may still contain errors, small amount of user interactions is necessary to keep our SSM under well control. 

By resorting to our SSM, we further propose a novel incremental learning framework to gradually incorporate unlabeled samples to enhance object detectors, as illustrated in Fig.~\ref{fig:architecture}.
%, the main components in our framework include: region proposal generation, high-consistency sample pseudo-labeling via our SSM, and low-consistency sample annotating via the AL.
%Though the cross image validation contributes to achieve high accuracy and efficiency of pseudo-labeling, the . In addition, the object detectors may misjudge some undefined object categories by the employed benchmarks from the external data. Retraining with these incorrect pseudo-annotations may result in detection performance drops. Hence, 
%By resorting to the proposed SSM process, we propose a novel incremental learning framework to gradually incorporate unlabeled samples to enhance the object detector without suffering from noisy samples/outliers. As illustrated in Fig.~\ref{fig:architecture}, the main components in our pipeline include: region proposal generation, cross image validation, high-consistency sample pseudo-labeling via the proposed SSM process, and low-consistency sample annotating via the AL. At the initial beginning, we first fine-tune ConvNet object detectors by a small set of annotated samples to obtain a not bad initialization. Then, we retrain the detectors in a one-off act by temporally pseudo-labeling the selected high-consistency region proposals from unlabeled or partially labeled images in every mini-batch. For the rest proposals, we intermittently annotate informative ones from them in an active user-query manner for the further training.
In our framework, inspired by the recently proposed techniques: Curriculum Learning (CL)~\cite{curriculun_learning} and Self-paced Learning (SPL)~\cite{spl_kumar}\cite{spcl}, we formulate the combining of the SSM and the AL as a concise optimization problem. Specifically, the SSM or AL process in our framework can jointly collaborate with each other. This is done by imposing a set of latent variables to progressively include samples into training. These variables determine whether a sample should be selected for pseudo-labeling or annotating. Meanwhile, the misleading of pseudo-labeled errors can be suppressed since the sample selection criterion is progressively optimized together with the batch-based incremental learning. In fact, the ambiguity of incorrectly annotated samples by users can also be eliminated, thanks to the correction of the majority pseudo-labeled samples. Hence, our SSM can further improve the robustness of classifiers against noisy samples/outliers in the pursuit of detection accuracy.

The \textbf{main contributions} of this work are two-fold. First, we propose a novel self-supervised process for automatically discovering and pseudo-labeling reliable region proposals via the cross image validation, which is compatible with the mini-batch based training for large-scale practical scenarios. Second, through fusing the proposed SSM process with the AL, we propose a principled framework with a concise optimization formulation and an alternative optimization algorithm.  Extensive experiments demonstrate that our proposed framework can not only achieve a clear performance gain by mining additional unlabeled data, but also outperform the dominantly state-of-the-art methods with significantly fewer annotations. %Therefore, our framework is capable of progressively boosting the object detectors by leveraging unlabeled or partially labeled data while minimizing user annotation.

%The rest of the paper is organized as follows. Section~\ref{sec:related_work} presents a brief review of related work. Section~\ref{sec:alg} overview the pipeline of our framework, followed by a discussion of model formulation and optimization. The experimental results, comparisons and component analysis are presented in Section~\ref{sec:exper}. Section~\ref{sec:conc} concludes the paper.

\section{Related Work}
\label{sec:related_work}
\textbf{Active Learning.} 
This branch of works mainly focuses on the sample selection strategy, i.e., how to pick out the most informative unlabeled samples for annotation. One of the most common strategies is the certainty-based selection \cite{lewis1994sequential,tong2002support}, in which the certainties are measured according to the prediction confidence on new unlabeled samples. The diversity of the selected instance over the unlabeled data has been also considered in~\cite{brinker2003incorporating}. Recently, Elhamifar {\em et al.}~\cite{con_AL} proposed to consider both the uncertainty and diversity measure via convex programming. Freytag {\em et al.}~\cite{Freytag14} presented a concept that generalizes previous methods based on the expected model change and incorporates the underlying data distribution. Vijayanarasimhan {\em et al.}~\cite{llal11CVPR} proposed a novel active learning approach in crowd-sourcing settings for live learning of object detectors, in which the system autonomously identifies the most uncertain instances via a hashing based solution. Rhee et al.~\cite{id17CSR} proposed to improve object detection performance by leveraging a collaborative sampling strategy, which integrates the uncertainty and diversity criteria from the AL and the feature similarity measurement of semi-supervised learning philosophy. However, these mentioned AL approaches usually emphasize those low-confidence samples (e.g., uncertain or diverse samples) while ignoring the rest majority of high-confidence samples. More recently, attempting to leverage these ignored samples, several works~\cite{aspl17TPAMI,ceal16tcsvt} have been proposed to progressively select the minority of most informative samples and pseudo-label the majority of high prediction confidence samples for network fine-tuning. Though these approaches have achieved promising performances, they have limitations due to that their defined hyper-parameters for pseudo-labeling is empirically set and updated within a single image context. Furthermore, these methods do not support mini-batch based training. Therefore, none of them has successfully proved their capability on handling large-scale object detection task. 

\textbf{Self-paced Learning.} Inspired by the cognitive principle of humans/animals, Bengio {\em et al.}~\cite{curriculun_learning} initialized the concept of curriculum learning (CL), in which a model is learned by gradually including samples into training from easy to complex. To make it more implementable, Kumar {\em et al.}~\cite{spl_kumar} substantially prompted this learning philosophy by formulating the CL principle as a concise optimization model named self-paced learning (SPL). Recently, several works~\cite{spcl,spl_reranking,spmf} provided more comprehensive understanding of the learning insight underlying CL/SPL, and formulated the learning model as a general optimization problem. 

Based on this framework, multiple SPL variants~\cite{spl_reranking,spld,spmf,spcl} have been proposed for object detection. Lee {\em et al.}~\cite{letf11CVPR} introduced a self-paced approach to focus on the easiest instances first, and progressively expands its repertoire to include more complex objects. Sangineto {\em et al.}~\cite{DBLP:journals/corr/SanginetoNCS16} presented a self-paced learning protocol for object detection that iteratively selects the most reliable images and boxes according to class-specific confidence levels and inter-classifier competitions. Dong {\em et al.}~\cite{fs17} proposed an object detection framework that uses only a few bounding box labels per category by consistently alternating between detector amelioration and reliable sample selection.

\textbf{Self-supervised Learning.} Aiming at training the feature representation without additional manual labeling, self-supervised learning (SSL) has first been introduced in \cite{Pal_computerrecognition} for vowel class recognition, and further extended for object extraction in ~\cite{Ghosh93self-organizationfor}. Recently, plenty of SSL methods\cite{MT_2017_ICCV,xl_2015_ICCV} have been proposed, e.g., Wang {\em et al.}~\cite{xl_2015_ICCV} proposed to employ visual tracking to provide the supervision for learning visual representations among thousands of unlabeled videos. Doersch {\em et al.}~\cite{MT_2017_ICCV} investigated multiple self-supervised methods to encourage the network to factorize the information in its representation. Different from these methods that focus on learning an optimal visual representation, our SSM intends to use the self-supervision to mine valuable information from unlabeled and partially labeled data.

\section{Self-supervised Sample Mining}
\label{sec:alg}
\subsection{Formulation}
In the context of object detection, suppose that we have $n$ region proposals from $m$ object categories. Denote the training sample set as $\mathcal{D}=\{{x}_{i}\}_{i=1}^{n} \subset R^{d}$, where ${x}_{i}$ is the $i$-th region proposal generated from the training images. We have $m$ detectors/classifiers (including the background) for recognizing each proposal by the one-vs-rest strategy. Correspondingly, we denote the label set of ${x}_i$ as $\mathbf{y}_i = \{y_{i}\}_{j=1}^m$, where $y_{i}^{(j)}$ corresponds to the label of ${x}_{i}$ for the $j$-th object category. i.e., if $y_{i}^{(j)}=1$, this means that ${x}_{i}$ is categorized as an instance of the $j$-th object category. We should give two necessary remarks on our problem setting. One is that most of sample labels $\mathbf{Y} = \{\mathbf{y}_{i}\}_{i=1}^n$ are unknown and need to be completed in the learning process. The initially annotated images are denoted by $\mathbf{I}$. The other remark is that the data $\{{x}_i\}_{i=1}^n$ might possibly be fed into the system in an incremental way. This means that the data scale might be consistently growing. %After a period of training, our SSM will be ameliorated based on the currently learned knowledge from data. 
The whole loss function for our proposed framework with SSM process is formulated as follows:
\begin{small}
\begin{equation}\label{eq:ssm}
Loss = \mathcal{L}_{loc}(\mathbf{W}) + \mathcal{L}_{cls}^\text{AL}(\mathbf{W}) + \mathcal{L}_{cls}^\text{SSM}(\mathbf{W}, \mathbf{V}),
\end{equation}\end{small}where $\mathcal{L}_{loc}(\mathbf{W})$ denotes the bounding box regression loss defined in~\cite{frcn}. $\mathcal{L}_{cls}^\text{AL}(\mathbf{W})$ and $\mathcal{L}_{cls}^\text{SSM}(\mathbf{W}, \mathbf{V})$ imply the classification loss for the AL and SSM processes, respectively. We define the AL process as $\mathcal{L}_{cls}^\text{AL}(\mathbf{W}) = \frac{1}{|\Omega_I|}\sum_{i \in \Omega_I} \sum_{j=1}^m \ell_{j}(x_i, \mathbf{W})$, where $\Omega_I$ denotes the labeled proposals from the currently annotated image $I$ ($I$ $\in \mathbf{I}$). $\ell_{j}(x_i, \mathbf{W})$ means the softmax loss of the proposal ${x}_{i}$ in the $j$-th classifier:
\begin{small}
\begin{equation*}
\begin{gathered}
\ell_{j}(x_i, \mathbf{W})=-\big(\frac{1 + y_{i}^{(j)}}{2}\log\phi_j(x_i; \mathbf{W}) + \\ \frac{1 - y_{i}^{(j)}}{2}\log(1-\phi_j(x_i; \mathbf{W}))\big), \\
\end{gathered}
\end{equation*} 
\end{small}where $\mathbf{W}$ represents the parameter of the CNN for all $m$ categories (including background), $\phi_j(x_i; \mathbf{W})$ denotes the probability of belonging to the $j$-th category for each region proposal $x_i$. 

To adaptively select $x_i$ for pseudo-labeling to update its $\mathbf{y}_i$, our SSM process introduces a set of latent weight variables, i.e., $\mathbf{V}=\{\mathbf{v}^{(j)}\}_{j=1}^{m}=\{[v_{1}^{(j)},\cdots,v_{n}^{(j)}]^{T}\}_{j=1}^m$, and is formulated as:
\begin{small}
\begin{equation}\label{eq:ssm}
\begin{gathered}
 \mathcal{L}_{cls}^\text{SSM}(\mathbf{W}, \mathbf{V}) = 
\frac{1}{|\overline{\Omega_I}|} \sum_{i \in \overline{\Omega_I}} \sum_{j=1}^{m} {v}_i^{(j)} \ell_{j}(x_i, \mathbf{W}) + R(x_i, v_i^{(j)}, \mathbf{W}) \\
 \text{s.t.} \quad \sum_{j=1}^m \vert y_i^{(j)} + 1 \vert \le 2, y_i^{(j)} \in \{-1, 1\},
\end{gathered} 
\end{equation}
\end{small}where $\overline{\Omega_I}$ denotes the unlabeled proposals from the unlabeled or partially labeled image $I$. The regularization function $R(\cdot)$ penalizes the sample weights linearly in terms of the loss. In this paper, we utilize the hard weighting regularizer due to its well adaptability to complex scenarios. The hard weighting regularizer is defined as:
\begin{small}
\begin{equation} \label{eq:r}
R(x_i, v_i^{(j)}, \mathbf{W}) = -f(x_i, \mathbf{W}) v_i^{(j)}.
\end{equation} 
\end{small}
Finally, we can directly calculate $v_i^{(j)}$ as:
\begin{small}
\begin{equation} \label{eq:v}
v_i^{(j)} = \left\{
\begin{aligned}
&1, & \ell_{j}(x_i, \mathbf{W}) \le f(x_i, \mathbf{W}), \\
&0, &otherwise. \\
\end{aligned}
\right.
\end{equation}
\end{small}

\begin{figure}[t]
\center
\includegraphics[width = 1 \columnwidth]{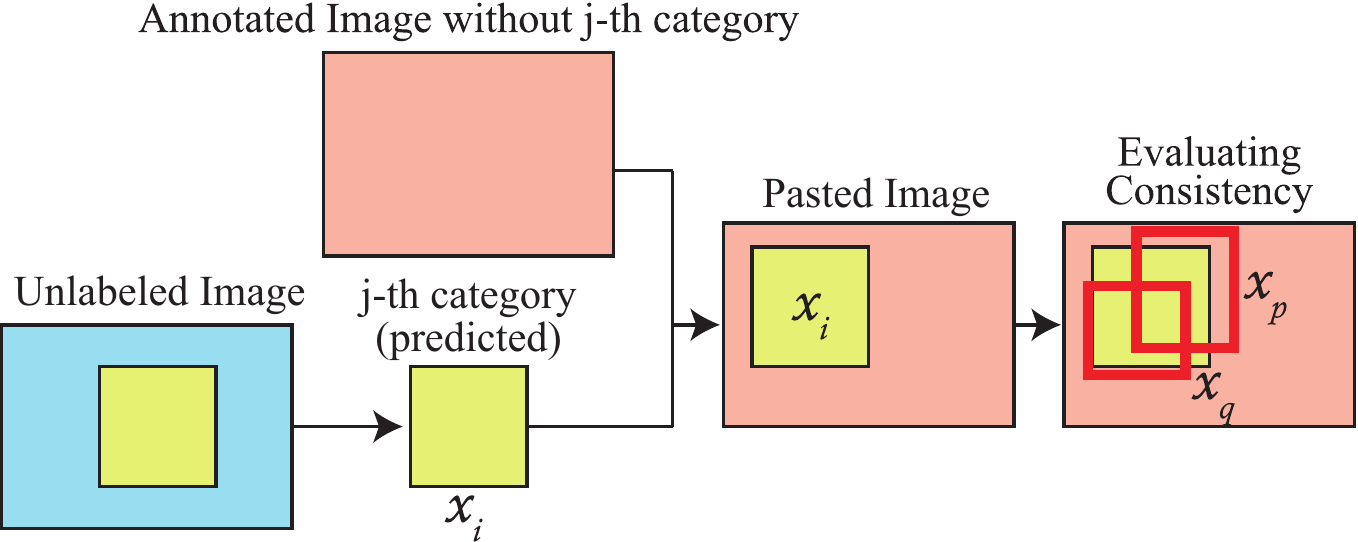}
\caption{The illustration of the proposed cross image validation. The proposal $x_i$ from the unlabeled image, predicted to belong to the $j$-th category, is randomly pasted into a certain annotated image without $j$-th category objects for consistency evaluation. The red bounding boxes $B_I(x_p)$ and $B_I(x_q)$ denote the newly generate region proposals $x_p$ and $x_q$ from the newly pasted image by the up-to-date classifier, respectively. As shown, the unlabeled and annotated images have entirely different context information (denoted in different color).}
\vspace{-8pt}
\label{fig:crossImgValidation}
\end{figure}

In contrast to the existing works~\cite{aspl17TPAMI,ceal16tcsvt} that rely on only an empirical hyper-parameter for each category to control the loss tolerance, we exploit a sample-dependent manner, i.e., the cross image validation $f(\cdot)$, to include samples into training. Therefore, our model can be considered as a self-supervised self-paced learning framework. As illustrated in Fig.~\ref{fig:crossImgValidation}, the cross image validation is regarded as the estimation consistency of reliable region proposals. Specifically, $f(\cdot)$ is defined as:
\begin{small}
\begin{equation}\label{eq:f}
\begin{gathered}
f(x_i, \mathbf{W}) = \\ \frac{\lambda}{|\Omega_I^{\overline{j}}|}  \sum_{p \in \Omega_I^{\overline{j}}} \textbf{1}\big(\text{IoU}\big(B_I(x_i), B_I(x_p)\big) \ge \gamma\big)  \phi_j(x_p; \mathbf{W}),
\end{gathered}
\end{equation} 
\end{small}where $\Omega_I^{\overline{j}}$ represents the labeled region proposals from the annotated image $I$ without $j$-th category objects for consistency evaluation. $\lambda$ denotes the pace parameter.
$B_I(x_i)$ denotes the bounding box of the proposal $x_i$ in the selected image $I$, while $\text{IoU}\big(B_I(x_i), B_I(x_p)\big)$ implies the intersection of union between two bounding boxes $B_I(x_i)$ and $B_I(x_p)$. Note that, $\gamma$ represents the threshold parameter to identify whether these two bounding boxes correspond to the same object, and it is set to 0.5 in all experiments according to the evaluation protocol of object detection. $\textbf{1}(\cdot)$ is the indicator function. If the proposal $x_p$ generated by the up-to-date detector from the newly pasted image includes the same object as $x_i$, i.e., $\text{IoU}\big(B_I(x_i), B_I(x_p)\big) \ge \gamma$, then we calculate its estimation consistency value $\phi_j(x_p; \mathbf{W})$, which denotes the possibility of $x_p$ being the $j$-th category during the cross image validation. 

\subsection{Alternative Optimization Strategy}
The alternative minimization is readily employed to solve this optimization. Specifically, the algorithm is designed by alternatively updating the sample importance weight set $\mathbf{V}$ via the SSM process, the label set $\mathbf{Y}$ via pseudo-labeling and the AL process, and the network parameters $\mathbf{W}$ via standard backpropagation. In the following, we introduce the details of these optimization steps. The convergence of this algorithm to the practical implementation of our framework will also be discussed.

\textbf{Updating $\mathbf{V}$}: Fixing the \{$\mathbf{Y}, \mathbf{X}, \mathbf{W}$\}, we can directly calculate $f(x_i, \mathbf{W})$ via Eqn.~(\ref{eq:f}), and further obtain $\mathbf{V}$ via Eqn.~(\ref{eq:v}). 

\textbf{Updating $\mathbf{Y}$}: After obtaining $\mathbf{V}$, we calculate the consistency score $s_i$ for sample selection as:
\begin{small}
\begin{equation} \label{eq:s}
\begin{gathered} 
j^* = \arg \max_{j\in [m]} \phi_j(x_p; \mathbf{W}), \\
s_i = \frac{1}{|\mathbf{I}|} \sum_{I \in \mathbf{I}} \frac{1}{|\Omega_I^{\overline{j^*}}|}
\sum_{p \in \Omega_I^{\overline{j^*}}} \phi_{j^*}(x_p; \mathbf{W}).
\end{gathered}
\end{equation}
\end{small}where $j^*$ represents the predicted category by the current detector with highest confidence.
%_I^{\overline{j}
%Then, Eqn.~(\ref{eq:f}) has a global optimum solution by computing the partial gradient equals with respect to Eqn.~(\ref{eq:r}). Considering $v_{p}^{(j)} \in [0,1]$, we have:
%\begin{small}
%\begin{equation}
%v_p^{(j)} = 1 + \log\phi_j(x_p; \mathbf{W}),
%\end{equation}
%\end{small}where $p \in \overline{\Omega^t_{j}}$. Through the same optimization manner, we can obtain $v_i^{(j)}$ in Eqn.~(\ref{eq:ssm}), i.e., $v_i^{(j)} = 1 - \ell_{ij}(x_i), i \not \in \Omega^t$. 
Note that, to reduce the computation cost, we only randomly pick out at most $N$ annotated images for evaluating the consistency of the proposal $x_i$. In all the experiments, we empirically set $N$ = 5 for the trade-off between the accuracy and efficiency. Given $\mathbf{S} = \{s_i\}_{i=1}^{n}$, we rank all unlabeled samples in a descending order for each classifier as~\cite{spl_reranking}, and pick out top-$k$ non-zero ones at most for each object category. $\mathcal{H}$ is regarded as high-consistency samples with assigned pseudo-labels. Specifically, these important samples for $m$ categories are defined as $\mathcal{H}=[H_1, ...,H_j,...,H_m] (|H_j| \le k)$, where $k$ is an empirical parameter to control the number of selected samples for each category.

Fixing \{$\mathbf{W}, \mathbf{V}, \{{x}_i\}_{i=1}^{\mathcal{H}}$\}, we optimize $\mathbf{y}_{i}$ of Eqn.~(\ref{eq:ssm}) which corresponds to solve the following problem for each high-consistency sample $i \in \mathcal{H}$ with its important weight vector $\mathbf{v}_i \not= \mathbf{0}$:
\begin{small}
\begin{equation} \label{eq:y}
\min_{\mathbf{y}_{i} \in \{-1, 1\}^m, i \in \mathcal{H}} \sum_{j=1}^{m} v_{i}^{(j)} \ell_{j}(x_i, \textbf{W}), \ \ \text{s.t.} \sum_{j=1}^{m} \vert y_{i}^{(j)} + 1 \vert \le 2, 
\end{equation}
\end{small}where $\mathbf{v}_i$ is fixed, and can be treated as constant.
The constraint $\sum_{j=1}^{m} \vert y_{i}^{(j)} + 1 \vert \le 2$ largely excludes all samples for pseudo-labeling except under the following two conditions: i) when $y_i^{(j)}$ is predicted to be positive by one classifier but all 
other classifiers produce negative predictions, or ii) when all classifiers predict $y_i^{(j)}$
to be negative, i.e., $x_i$ 
is rejected by all classifiers and identified as belonging to an undefined object category. These are the rational cases for practical object detection in large-scale scenarios. Note that we optimize $\mathbf{Y}$ by exhaustively attempting to assign -1 or 1 to each sample for all $m$ categories to minimize Eqn.~(\ref{eq:y}). The computational cost of this process is acceptable because we only need to take  $m+1$ attempts. Through this fashion,  Eqn.~(\ref{eq:y}) always has a clear solution by enforcing pseudo-labels on those top-ranked high-consistency sample set $\mathcal{H}$. This is exactly the mechanism underlying a re-ranking technique~\cite{spl_reranking}. Compared with the previous methods~\cite{ceal16tcsvt,aspl17TPAMI}, our framework can effectively suppress the error accumulation during the incremental pseudo-labeling via the following two advantages: (i) The cross image validation can provide more accurate and robust estimations under various challenging image contexts; (ii) All the pseudo-labels are disposable. They will be discarded after each mini-batch iteration. These advantages are beneficial for the detector to avoid being misled by the accumulate errors.

%Note that actually only those high-consistency samples with positive weights, as calculated in the last updating step for $\mathbf{v}$, are meaningful for the solution. This implies the physical interpretation for this optimization step: we iteratively find the high-consistency samples based on the current classifier, and then  

%%%%%%%%%%%%%%%%%%%%%%%%%%%%%%%%%%%%%%%%%%%%%%%%%%%%%%%%%%%%%%%%%%%%%%%%%%%%%%%%%%%%%%%%%%%%%%%%%%%

\textbf{Low-consistency Sample Annotating}: After pseudo-labeling high-consistency samples in such a self-supervised manner, we employ the AL process to update the annotated image set $\mathbf{I}$ by providing more informative guidance based on human knowledge. The AL process aims to select most informative unlabeled samples and to label them as either positive or negative by requesting user annotation. Our selection criteria are based on the classical uncertainty-based strategy \cite{lewis1994sequential,tong2002support}. Specifically, we collect those samples with quite small $s_i$ after performing the cross image validation. Then, we utilize the current classifiers to predict their labels. Those predicted as more than two positive labels (i.e., predicted as the corresponding object category) actually represent these samples making the current classifiers ambiguous. We thus adopt them as so called ``low-consistency'' samples and randomly add $z$ of them into low-consistency sample set $\mathcal{U}$ for further manually annotation by active users. Actually, other similar criterion can be utilized in this step. We employed this simple strategy just due to its intuitive rationality and efficiency.

\textbf{Updating $\mathbf{W}$}: Fixing $\{\mathcal{D},\mathbf{V}, \mathbf{Y}\}$, the original model in Eqn.~(\ref{eq:ssm}) can be approximated as:
\begin{small}
\begin{equation}\label{eq:w}
\begin{gathered}
\min_{\mathbf{W}} \frac{1}{|\mathcal{H} \cup \{\Omega_I\}_{I \in \mathbf{I}}|}
\sum_{i \in \mathcal{H} \cup \{\Omega_I\}_{I \in \mathbf{I}}} \sum_{j=1}^{m} \ell_{j} (x_i, \mathbf{W}) + \mathcal{L}_{loc}(\mathbf{W}). \\
\end{gathered} 
\end{equation}
\end{small}Thus, we can update the network parameters $\mathbf{W}$ by standard backpropagation. Note that, we do not consider the regularization function $R(\cdot)$, which is introduced to regularize the latent variable set $\mathbf{V}$.

\begin{algorithm}[t]
\caption{Alternative Optimization Strategy}
\label{alg:alg_overview}
\begin{algorithmic}[1]
\REQUIRE Input dataset $\{{x}_{i}\}_{i=1}^{n}$
\ENSURE Output model parameters $\{\mathbf{W}\}$
\STATE Initialize network parameters $\mathbf{W}$ with pre-trained CNN, initially annotated samples $\mathbf{I}$, sample weight set $\mathbf{V}$ and the corresponding consistency score set $\mathbf{S}$;
\STATE
\textbf{while} true \textbf{do} \\
\STATE \ \ \ \ \textbf{for all} mini-batch = $1,...,T$ \textbf{do}
\STATE \ \ \ \ \ \ \ Update $\mathbf{V}$ and $\mathbf{S}$ by the SSM process via Eqn.~(\ref{eq:v}) and Eqn.~(\ref{eq:s}), respectively;
\STATE \ \ \ \ \ \ \ Update $\mathcal{H}$ by the re-ranking;
\STATE \ \ \ \ \ \ \ Update $\{\mathbf{y}_i\}_{i \in \mathcal{H}}$ by pseudo-labeling via Eqn.~(\ref{eq:y});
\STATE \ \ \ \ \ \ \ Update $\mathbf{W}$ by standard backpropagation Eqn.(\ref{eq:w});
\STATE \ \ \ \ \textbf{end for}
\STATE \ \ \ \ Update low-consistency sample set $\mathcal{U}$;
\STATE \ \ \ \ \textbf{if} $\mathcal{U}$ is not empty \textbf{do}
\STATE \ \ \ \ \ \ \ Update the annotated region proposals $\{\Omega_I\}_{I \in \mathbf{I}}$ with $\{\mathbf{y}_i\}_{i \in \mathcal{U}}$ by the AL;					 
\STATE \ \ \ \ \textbf{else}
\STATE \ \ \ \ \ \ \ \textbf{break};
\STATE \ \ \ \ \textbf{end if}
\STATE\textbf{end while}
\RETURN $\mathbf{W}$;
\end{algorithmic}
\end{algorithm}

The entire algorithm can then be summarized into Algorithm~\ref{alg:alg_overview}. It is easy to see that this algorithm finely accords with the pipeline of our proposed framework in Fig.~\ref{fig:architecture}.

\textbf{Convergence Analysis:} Our framework can guarantee the convergence to a local optimum based on its implementation. The reason is three-fold: (i) the objective function Eqn.~(\ref{eq:ssm}) w.r.t $\mathbf{V}$ is convex; (ii) network fine-tuning in Eqn.~(\ref{eq:w}) via backpropagation can converge to a local optimal; (iii) as the model becomes mature, the AL process stops when no low-consistency samples are found.

\section{Experiments}
\label{sec:exper}

To justify the effectiveness of our proposed SSM process and framework, we have conducted a number of experiments on the public VOC 2007/2012 benchmarks~\cite{voc2007}, whose data are usually divided into two-fold: trainval and test. To evaluate the model performance on the VOC 2007 benchmark, we regard the VOC 2007 trainval as the initial annotated samples, and consider the VOC 2012 trainval data as unlabeled data that need to be mined. Therefore, the active user annotating process equals to fetch the VOC 2012 annotations. As for the VOC 2012 benchmark, we employ the VOC 2007 trainval and test set for initialization. Moreover, we regard the large-scale object detection dataset COCO~\cite{coco14ECCV} as the `secondary' unlabeled data. In other words, we will perform sample mining on it only when all the VOC 2012 trainval annotations have been used. As for the annotation key `annotated' and `pseudo', the first one represents the proportion of the user annotations appended/fetched during the training over the pre-given annotations (i.e., the VOC 2007 trainval), which are used for initializing the object detectors. `pseudo' implies the percentage of pseudo-labeled object proposals from unlabeled data over the pre-given annotations. The lower `annotated' value is, the less user efforts for annotating are required. The higher `pseudo' value is, the more pseudo-labeled samples are obtained. Hence, when achieving the same performance, a superior method should have lower `annotated' but higher `pseudo' values. 

We adopt the PASCAL Challenge protocol, i.e., a correct detection should has more than 0.5 IoU with the ground truth bounding-box, and exploit the mean Average Precision (mAP) as for the evaluation metric. In all experiments, we set parameters \{$\lambda$, $k$, $N$, $\gamma$, $z$\} = \{0.9, 500, 5, 0.5, 100\}.  The fine-tuning manner for the RFCN pipeline is the same as \cite{rfcn16NIPS}, except that we treat the COCO trainval set as unlabeled data for mining rather than pre-train the network. In the testing phase, we use a single scale of 600 pixels as input except for the VOC 2012 test set, which we use a multi-scale test as ~\cite{spp15PAMI}. All the experiments are conducted on a desktop with Intel CPU 3.4GHz and four Titan Xp GPUs. The testing time of our framework is 120 millisecond/image, and our training time is 620 millisecond/image. As for the base line method (i.e., RFCN), its testing time is 120 millisecond/image and training time is 150 millisecond/image. 

%Specifically, we employ multi-scale training (400, 500, 600, 700, 800) to update the network parameters of RFCN by using a weight decay of 0.0005 and a momentum 0.9. The model, including the region proposal network (RPN), is trained with 4 GPUs (mini-batch size = 4) via a learning rate of 0.001.

In order to demonstrate that our proposed framework is general to different network architecture and object recognition framework, we have incorporated our framework into the Fast R-CNN (FRCN) pipeline with AlexNet~\cite{alexnet12NIPS} and the new state-of-the-art RFCN~\cite{rfcn16NIPS} with ResNet-101~\cite{He_2016_CVPR}. We denote these variants as ``FRCN+Ours'' and ``RFCN+Ours'', respectively. To validate the superior performance of the proposed framework, we have compared it with the CEAL~\cite{ceal16tcsvt} and K-EM~\cite{em17} approaches. Note that, since the CEAL is designed for image classification and is not mini-batch friendly, we extended it for object detection by alternatively performing sample selection and fine-tuning the CNN. Since the source code of K-EM~\cite{em17} is not publicly available, we have obtained the results from their paper~\cite{em17} directly. We denote these methods as ``FRCN+CEAL,'' and ``FRCN+K-EM'', respectively. Moreover, we have also included five baseline methods that ignore the pseudo-labeling of unlabeled samples: ``FRCN+RAND'', which randomly selects region proposals for user annotations,  FRCN+EMC (Expected-Model-Change)~\cite{Freytag14}, FRCN+Entropy~\cite{survey}, FRCN+ELC (Expected-Labeling-Change)~\cite{label} and FRCN+Density~\cite{survey}. For a fair comparison, we initialize all the methods by providing only 5\% annotations, and allow FRCN+Ours and FRCN+CEAL to mine unlabeled samples only from the VOC 2007 train/val set. Moreover, we repeat the testing by five trails to report the average mAP with standard variance to present a comprehensive evaluation. 

\begin{figure}[tbp]
\center
\includegraphics[width=0.65 \columnwidth]{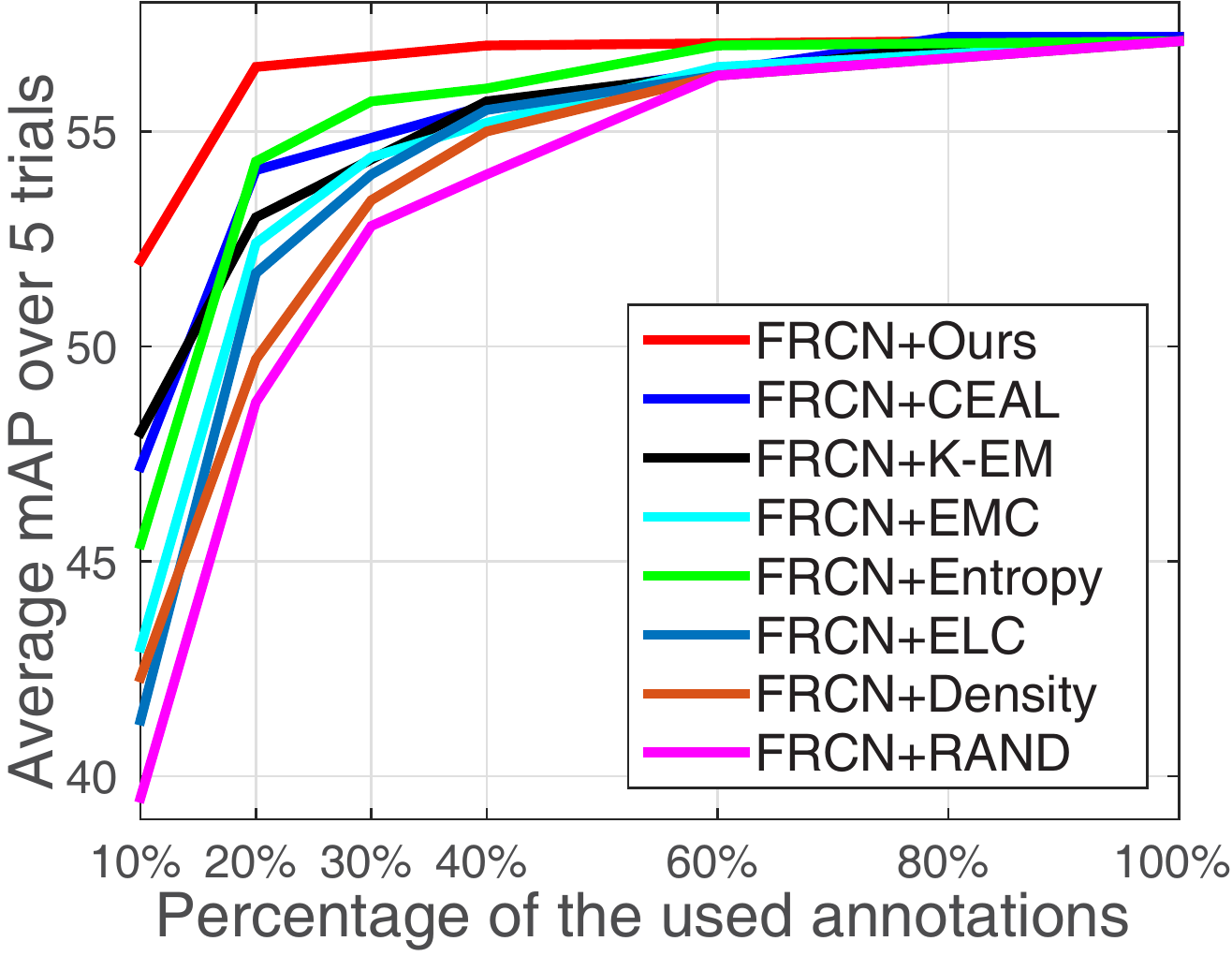}
%\vspace{-10pt}
\caption{Quantitative comparison of detection performance (mAP) on the PASCAL VOC 2007 test set}\label{fig:aoa}
\vspace{-15pt}
\end{figure}

\begin{table}[tbp]
\center
%\vspace{-5pt}
\footnotesize
\center
\caption{Test set mAP for VOC 2007/2012 under the RFCN~\cite{rfcn16NIPS} pipeline. The entries with the best APs with each sub-table are bold-faced. Annotation key: `annotated' denotes the proportion of the user annotations appended/fetched from the VOC 2012 trainval set during the training over the initial annotations (i.e., 07 denotes the VOC 2007 trainval set, while 07+ denotes the VOC 2007 trainval and test sets), which are used for initializing the object detectors; `pseudo' implies the percentage of pseudo-labeled object proposals from unlabeled data (i.e., the VOC 2012/COCO trainval set) over the pre-given annotations. }\label{tab:07mAP12}\label{tab:12mAP07}\label{tab:07mAP12_2}\label{tab:12mAP07_2}
\setlength{\tabcolsep}{2.5pt}
\begin{tabular}{c|c | c c | c  c | c }
\hline
\hline
& Method & initial & test & annotated & pseudo & mAP  \\
\hline
\multirow{10}*{(a)} & RFCN & 07 & 07 & 0\% & 0\% & 73.9  \\
& RFCN+RAND & 07 & 07& 20\% & 0\%  & 75.6$\pm$1.0  \\
& RFCN+RAND & 07 & 07& 60\% & 0\% & 76.5$\pm$1.1 \\
& RFCN+RAND & 07 & 07& 100\% & 0\% & 77.2$\pm$0.9 \\
& RFCN+RAND & 07 & 07& 200\% & 0\% & 79.1$\pm$0.4 \\
& RFCN+Ours & 07 & 07& 20\% & 300\%  & 76.0$\pm$0.1 \\
& RFCN+Ours & 07 & 07& 60\% & 400\% & 77.4$\pm$0.2 \\
& RFCN+Ours & 07 & 07& 100\% & 500\% & 78.3$\pm$0.2 \\
& RFCN+Ours & 07 & 07& 200\% & 800\% & 79.7$\pm$0.2 \\
& RFCN+Ours & 07 & 07& 200\% & 1000\% &\bf 80.6$\pm$0.2 \\
\hline
\hline
\multirow{9}*{(b)} & RFCN & 07+ & 12 & 0\% & 0\% &  69.1 \\
& RFCN+RAND & 07+ & 12 & 10\%  & 0\% & 71.5$\pm$1.1  \\
& RFCN+RAND & 07+ & 12 & 30\% & 0\% & 72.7$\pm$1.3  \\
& RFCN+RAND & 07+ & 12& 50\% & 0\% & 74.4$\pm$1.0  \\
& RFCN+RAND & 07+ & 12 & 100\% & 0\% & 76.8$\pm$0.4  \\
& RFCN+Ours & 07+ & 12 & 10\%  & 100\% & 72.6$\pm$0.1  \\
& RFCN+Ours & 07+ & 12 & 30\% & 150\% & 73.6$\pm$0.1 \\
& RFCN+Ours & 07+ & 12 & 50\% & 200\% & 75.5$\pm$0.2 \\
& RFCN+Ours & 07+ & 12 & 100\% & 200\% & 77.3$\pm$0.2 \\
& RFCN+Ours & 07+ & 12 & 100\% & 800\% &\bf 78.1$\pm$0.2 \\
\hline
\hline
\multirow{8}*{(c)} & RFCN & 07 & 07& 0\% & 0\% & 73.9	 \\
& RFCN+SPL & 07 & 07& 0\% & 300\% & 74.1$\pm$0.5  \\
& RFCN+SPL & 07 & 07& 0\% & 400\% & 74.7$\pm$0.6  \\
& RFCN+SSM & 07 & 07& 0\% & 300\%& 75.6$\pm$0.2  \\
& RFCN+SSM & 07 & 07& 0\% & 400\%& 76.7$\pm$0.3  \\
& RFCN+AL  & 07 & 07& 20\% & 0\% & 75.5$\pm$0.1 \\
& RFCN+AL  & 07 & 07& 60\% & 0\% & 77.0$\pm$0.2 \\
& RFCN+AL  & 07 & 07& 100\% & 0\% &\bf 77.5$\pm$0.2 \\
\hline
\hline
\multirow{6}*{(d)}  & RFCN     & 07+ & 12& 0\% & 0\% &  69.1  \\
& RFCN+SPL & 07+ & 12& 0\%  & 100\%& 70.9$\pm$0.5   \\
& RFCN+SSM & 07+ & 12& 0\% & 100\%& 72.1$\pm$0.3  \\ 
& RFCN+AL  & 07+ & 12& 10\% & 0\% & 71.8$\pm$0.1   \\
& RFCN+AL  & 07+ & 12& 30\%  & 0\% &  73.0$\pm$0.2   \\
& RFCN+AL  & 07+ & 12& 50\% & 0\% &\bf 74.7$\pm$0.2  \\
\hline
\hline
\end{tabular}
\vspace{-15pt}
\end{table}

\subsection{Results and Comparisons}
The Fig.~\ref{fig:aoa} demonstrates the comparison of detection performance using the Fast-RCNN (FRCN) pipeline with AlexNet~\cite{alexnet12NIPS} on the VOC 2007 test set. As illustrated in Fig.~\ref{fig:aoa}, our proposed framework FRCN+Ours consistently outperforms all the compared methods by clear margins under all annotation settings. Specifically, FRCN+Ours can achieve the equivalent of a fully supervised performance (i.e., FRCN with 100\% user annotations) when fetching only approximately 30\% user annotations, while most of the compared methods require nearly 60\%. These results indicate the superior performance of our framework.

To demonstrate the feasibility and great potential of our framework, we have conducted amount of experiments to fine-tune RFCN with ResNet-101 (well pre-trained on ImageNet) on the VOC 2007/2012 trainval set by using our framework and the compared baseline RFCN+RAND. The compared results on the VOC 2007 and 2012 benchmarks are illustrated in Tab.~\ref{tab:07mAP12} (a)(b), respectively. By controlling the number of training iterations, RFCN+Ours and RFCN+RAND can fetch different amounts of annotations from the VOC 2012 trainval set.

\begin{figure*}[t]
\centering
\includegraphics[width = 0.95\textwidth]{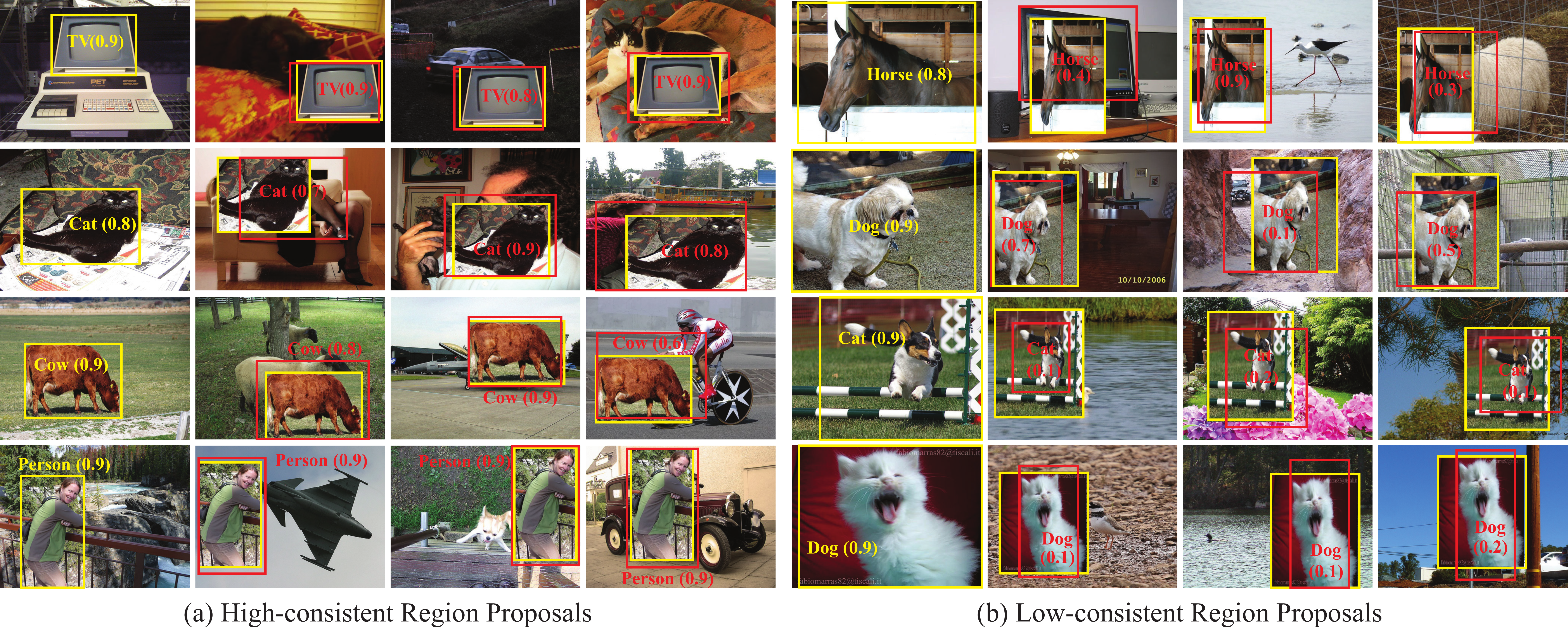}
\caption{Selected (a) high-consistent and (b) low-consistent region proposals with pseudo-labels in yellow via cross image validation. The first column lists the region proposal with predicted label and high prediction confidence from the unlabeled images of the PASCAL VOC 2012 dataset. The region proposal is randomly put into the validation images for object detection. The predicted bounding box that has more than 0.5 IoU with this proposal is illustrated in red. The corresponding prediction confidence (ranging from 0 to 1) for being the pseudo-labeled category is also in red. It is obvious from (a) that those high-consistent proposals are of high quality for network fine-tuning. As one can see from (b), the confidence of those bounding boxes on the validation images is low-consistent due to inaccurate bounding box (first two rows) or incorrect pseudo-labels (last two rows)}\label{fig:example}
\vspace{-12pt}
\end{figure*}

As one can see from Tab.~\ref{tab:07mAP12} (a)(b), both RFCN+RAND and RFCN+Ours gradually obtain increased detection accuracy when the number of annotations increases. Our framework consistently outperforms the compared baseline RFCN+RAND under all appending conditions on both the VOC 2007 and 2012 benchmarks by a clear margin. Moreover, our framework surpasses RFCN+RAND by nearly 1.5\% (80.6\% vs 79.1\%) on the VOC 2007 benchmark and 1.3\% (78.1\% vs 76.8\%) on the VOC 2012 benchmark with significantly small variations when sufficient pseudo-labeled object proposals are provided. This validates the effectiveness of our framework. Some examples of the selected high-consistency and low-consistency region proposals via the cross image validation are depicted in Fig.~\ref{fig:example}.

\begin{figure}[tbp]
\center
\includegraphics[width=0.65 \columnwidth]{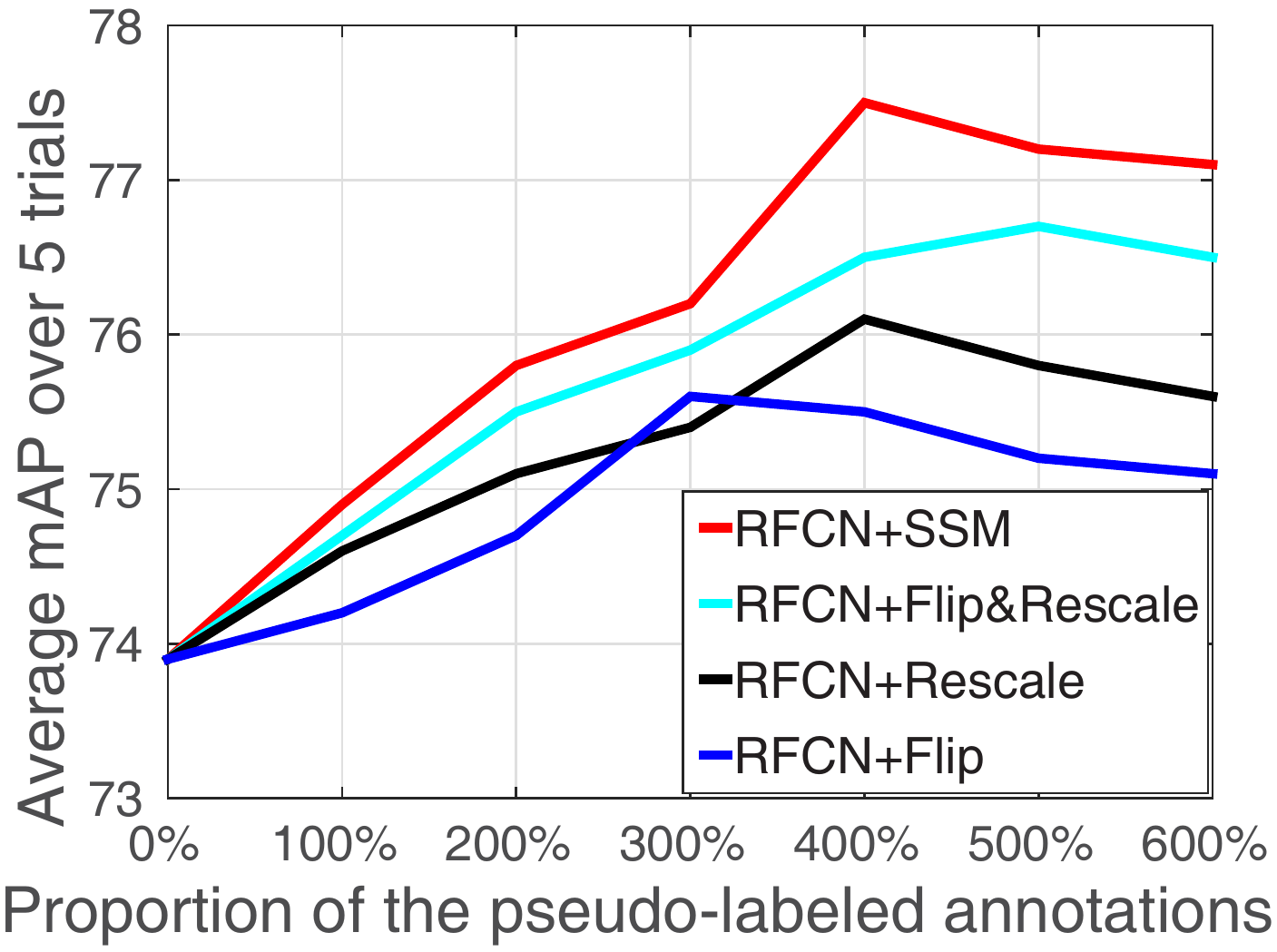}
%\vspace{-10pt}
\caption{Quantitative comparison of average mAP on the VOC 2007 test set}\label{fig:aoa2}
\vspace{-12pt}
\end{figure}

\subsection{Ablation Study}
To validate the contribution of each component inside our framework, we have also conducted sufficient experiments for empirical analysis. The variant of our framework that discarding the active learning process is denoted as ``FRCN+SSM'' / ``RFCN+SSM''. Similarly, these pipelines with Active Learning (AL), Self-paced Learning (SPL) are denoted by ``RFCN+AL'', ``RFCN+SPL'', respectively. RFCN+AL adaptively collects low-confidence proposals to request the annotations and stops when no low-confidence samples are found. RFCN+SPL is implemented according to \cite{ceal16tcsvt}.

As illustrated in Tab.~\ref{tab:07mAP12_2} (c)(d), given the same amount of annotations during initialization, RFCN+SSM performs significantly better than RFCN+SPL on both VOC 2007 and the VOC 2012 test set. Specifically, RFCN+SSM achieves a nearly 2\% performance improvement (76.7\% vs 74.7\%) with small variations over RFCN+SPL by pseudo-labeling the same amount of high-consistency region proposals for training on the VOC 2007 benchmark. A consistent performance gain of approximately 1.2\% (72.1\% vs 70.9\%) is obtained on the VOC 2012 benchmark by RFCN+SSM. These results validate the significant contribution of the proposed SSM process on mining reliable region proposals for improving object detection.

We have also compared our self-supervised sample mining (SSM) process with three baseline methods under the AL process disabled setting. RFCN+Flip implies horizontally flipping images for validation, while RFCN+Rescale represents randomly rescaling an image from 50\% to 200\% of its original size. RFCN+Flip\&Rescale means the fusion of them. As shown in Fig.~\ref{fig:aoa2}, RFCN+SSM consistently outperforms all the competing methods by clear margins at all pseudo-labeling proportions. Moreover, compared to these baselines, RFCN+SSM obtains a slighter performance drop (caused by the accumulated pseudo-labeling errors) after reaching its peak performance. This also proves the effectiveness of our SSM for suppressing the error accumulation.

Tab.~\ref{tab:07mAP12_2} (c)(d) also demonstrates that RFCN+AL consistently outperforms the baseline RFCN and RFCN+SSM. Though the improvements are minor compared to the best performance of the RFCN+SSM, our proposed AL stage is still beneficial for promoting object detection. This slight improvement occurs because the informative samples with great potential for improving performance follow a long-tail distribution, as reported in~\cite{Wang_2017_CVPR}. Therefore, it is necessary to employ abundant training samples by asking active users to provide labels or finding other assistance. Fortunately, our proposed high-consistency sample pseudo-labeling via the SSM process is an effective way to address this issue.

The results of weaker initialization (5\% and 10\% annotations from the VOC 2007 train/val set) are listed in Tab.~\ref{tab:ss}. As shown, our RFCN+SSM achieves a consistent performance gain of about 2\% over the original RFCN. Compared to RFCN+RAND, RFCN+Ours obtains about 1.5\% higher average mAP with much lower variances. However, compared to RFCN+Ours, FRCN+Ours in Fig.~\ref{fig:aoa} obtains a higher mAP gain. The reason is FRCN and RFCN use different algorithms (Selective Search~\cite{ss11ICCV} vs. Region Proposal Network~\cite{frcn}) to generate object proposals. Since our objective is to mine samples from these proposals, our model obtains various performance boosts based on the quality of proposals. This shows the generality and effectiveness of our model over different proposals.

\section*{Acknowledgment}
This work was supported in part by the Hong Kong Polytechnic University’s Joint Supervision Scheme with the Chinese Mainland, Taiwan and Macao Universities (Grant no. G-SB20). This work was also supported by HK RGC General Research Fund (PolyU 152135/16E), in part by Guangdong ``Climbing Program'' Special Funds under Grant pdjhb0010, in part by National Natural Science Foundation of China (NSFC) under Grant U1611461 and Grant 61702565, and in part by Science and Technology Planning Project of Guangdong Province of No.2017B010116001.

\begin{table}[t]
\footnotesize
\center
\setlength{\tabcolsep}{1.5pt}
\caption{Test set mAP for VOC 2007 under the RFCN pipeline with ResNet-101.}
\label{tab:ss}
\begin{tabular}{c | c c | c || c c | c}
\hline
\hline
Method & initial & annotated & mAP & initial & annotated & mAP  \\
\hline
RFCN & 5\% & 0\% & 49.4$\pm$1.4 & 10\% & 0\% & 56.8$\pm$1.3 \\
RFCN+SSM & 5\% & 0\% & 51.6$\pm$0.3 & 10\% & 0\% & 59.4$\pm$0.2 \\
\hline
RFCN+RAND & 5\% & 30\% & 53.3$\pm$1.1 & 10\% & 30\% & 60.4$\pm$1.5 \\
RFCN+Ours & 5\% & 30\% & 55.1$\pm$0.1 & 10\% & 30\% & 62.9$\pm$0.2 \\
\hline
\hline
\end{tabular}
% ``initial'' denotes the proportion of user annotations for initialization. ``annotated'' denotes the proportion of user annotations appended/fetched from VOC 2007 trainval set.}
\vspace{-15pt}
\end{table}

\section{Conclusion}
\label{sec:conc}
In this paper, we have introduced a principled Self-supervised Sample Mining (SSM) process, and justified its effectiveness in mining valuable information from unlabeled or partially labeled data to boost object detection. We further involve this process in the AL pipeline with a concise formulation, which is developed for retraining object detectors via faithfully pseudo-labeled high-consistency object proposals after our proposed cross image validation. The proposed SSM process contributes to effectively improve the detection accuracy and the robustness against noisy samples. Meanwhile, the rest samples, being low consistency (high uncertainty) by the current detectors, can be handled by the AL, which benefits to generate reliable and diverse samples gradually. In the future, we will apply our SSM to improve other specific visual detection task with unlabeled web images/videos.

{\small
\bibliographystyle{ieee}
\bibliography{aspl}
}

\end{document}